# Tracking Live Fish from Low-Contrast and Low-Frame-Rate Stereo Videos

Meng-Che Chuang, Jenq-Neng Hwang, Kresimir Williams, and Richard Towler

*Abstract*—**Non-extractive fish abundance estimation with the aid of visual analysis has drawn increasing attention. Unstable illumination, ubiquitous noise and low frame rate video capturing in the underwater environment, however, make conventional tracking methods unreliable. In this paper, we present a multiple fish tracking system for low-contrast and low-frame-rate stereo videos with the use of a trawl-based underwater camera system. An automatic fish segmentation algorithm overcomes the low-contrast issues by adopting a histogram backprojection approach on double local-thresholded images to ensure an accurate segmentation on the fish shape boundaries. Built upon a reliable feature-based object matching method, a multiple-target tracking algorithm via a modified Viterbi data association is proposed to overcome the poor motion continuity and frequent entrance/exit of fish targets under low-frame-rate scenarios. In addition, a computationally efficient block-matching approach performs successful stereo matching, which enables an automatic fish-body tail compensation to greatly reduce segmentation error and allows for an accurate fish length measurement. Experimental results show that an effective and reliable tracking performance for multiple live fish with underwater stereo cameras is achieved.**

*Index Terms*—**fish abundance estimation, low frame rate video, multiple target tracking, stereo imaging, underwater video**

## I. INTRODUCTION

FISH abundance estimation [1], which often calls for the use of bottom and midwater trawls, is critically required for the commercially important fish populations in oceanography and fisheries science. However, fish captured by trawls often do not survive, and thus trawl survey methods are inappropriate in some areas where fish stocks are severely depleted. To address these needs, we developed the Cam-trawl [2] to conduct video-based surveys. The absence of the codend allows fish to pass unharmed to the environment after being sampled (captured by cameras). The captured video data provide much of the information that is typically collected from fish that are retained by traditional trawl methods.

Video-based sampling for fish abundance estimates generates vast amounts of data, which present challenges to data analyses. These challenges can be reduced by using video processing techniques for automated detection, segmentation, tracking, length/size measurement and recognition. A successful development of these algorithms will greatly ease one of the most onerous steps in video-based sampling. Specifically, object tracking provides a mean of avoiding double counting of individual fish that are captured in multiple frames, and allows for more accurate length estimation by averaging several measurements of the same fish.

Underwater video processing for fish detection, tracking and counting using monocular or stereo cameras have been investigated in [3]–[8], [27]. There are, however, several challenges for underwater image/video analyses. First, the fast attenuation and non-uniformity of LED illumination make many foreground objects have relatively low contrast with the background, and fish with similar ranges from the cameras can have significantly different intensity because of the differences in angle of incidence as well as reflectivity of fish body among species. These factors make segmentation of fish difficult. Second, the ubiquitous noise is created by non-fish objects such as bubbles, organic debris and invertebrates, which can easily be mistaken as real fish. Third, low frame rate (LFR) of capturing results in poor motion continuity and frequent entrance/exit of the field of view for fish targets (see Fig. 1 (b)), and thus makes conventional multiple-target tracking algorithms [7], [8] infeasible under such circumstances.

On the other hand, automatic stereo matching (or correspondence) has been one of the most heavily investigated topics in computer vision [9]–[12], [30]. Nevertheless, most state-of-the-art stereo matching approaches suffer seriously from their intensive computations, making them rather infeasible for a real-time framework.

All these issues motivate us to resort to a novel solution to underwater live fish length measurement and tracking for video-based fishery survey systems. In this paper, a multiple fish tracking algorithm for trawl-based underwater camera systems is proposed to perform automatic fish size estimation and counting. We have overcome the difficulties imposed by uncontrolled illumination and noisy video capturing, which are very common in the underwater scenarios. In addition, stereo correspondence of objects is exploited for not only being

Manuscript received November 8, 2013; revised March 19 and June 23, 2014; accepted September 4, 2014. This work was supported by the National Marine Fisheries Service's Advanced Sampling Technology Working Group, National Oceanic and Atmospheric Administration, Seattle, WA, USA. This paper was recommended by Associate Editor Dr. Peng Yin.

M.-C. Chuang and J.-N. Hwang are with the Department of Electrical Engineering, University of Washington, Seattle, WA 98195 USA (e-mail: {mengche, hwang}@uw.edu).

K. Williams and R. Towler are with the Alaska Fisheries Science Center, National Oceanic and Atmospheric Administration (NOAA), Seattle, WA 98115 USA (e-mail: {kresimir.williams, rick.towler}@noaa.gov).

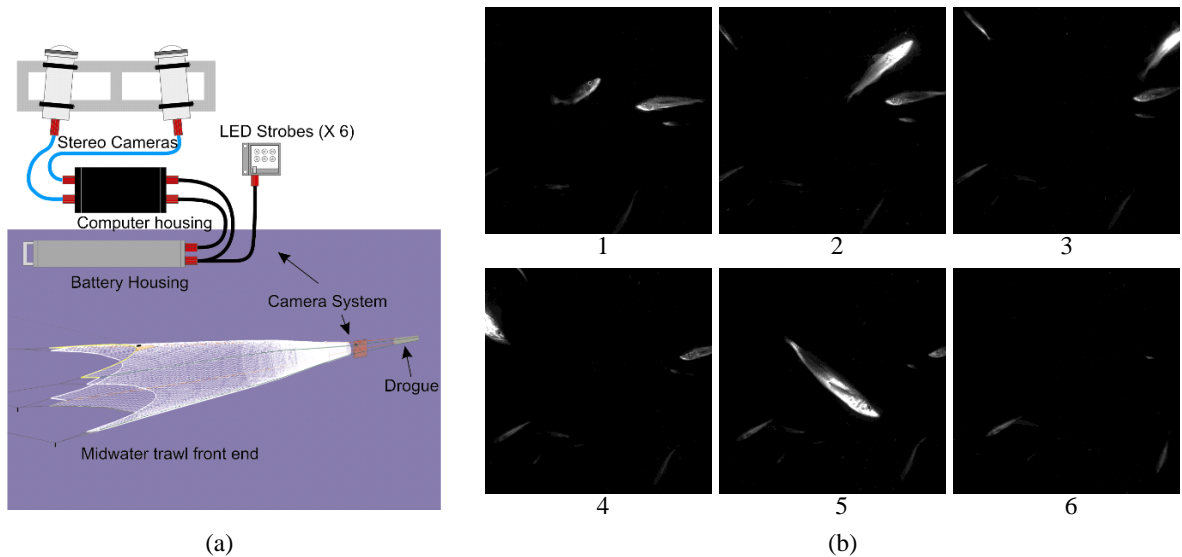

1  2  3

4  5  6

(a)  (b)

Fig. 1. (a) Illustration of the Cam-trawl underwater fish imaging system and (b) underwater video captured at 5 frames per second, showing the abrupt motion and frequent entrance/exit of fish.

incorporated with dynamic programming to establish a low-frame-rate video target tracker but also giving a reliable length measurement in 3-D space. The contributions of this paper include an adaptive object segmentation algorithm that overcomes the challenges imposed by low contrast and uneven illumination by modifying Otsu's thresholding method [21] and histogram backprojection procedure [14], and a novel multiple-target tracking algorithm to track fish with abrupt movement due to low frame rate by developing a feature-based temporal matching approach and extending the Viterbi data association used in single-target tracking. With the stereo vision available, we also proposed a fast and effective stereo matching approach followed by a self-compensation scheme to accomplish the fish-pair matching and thus allows for length estimation of each fish target.

The remaining of this paper is organized as follows. Section II gives an overview of the configuration and functional modules of our Cam-trawl for underwater video capturing. In Section III, the automatic fish segmentation algorithm is described. Section IV introduces the multiple-fish tracking algorithm for low-frame-rate video, as well as the fish length measurement approach using stereo vision techniques. Section V demonstrates our experimental results. Section VI provides discussions on the proposed system in several aspects. Finally, a conclusion is given in Section VI.

## II. CAMERA SYSTEM OVERVIEW

The Cam-trawl [2] represents a new class of mid-water imaging sampler to study the marine environment. With ongoing development, however, the Cam-trawl is poised to become a standard marine surveying tool to provide a more holistic view of the marine environment, and improve the management of our marine resources.

As shown in Fig. 1 (a), the Cam-trawl consists of two high-resolution machine vision cameras, a series of LED strobes, a computer, microcontroller, sensors, and battery power supply. The cameras and battery pack are housed in separate 4-inch diameter titanium pressure housings, and the computer, microcontroller and sensors are placed in a single 6-inch diameter aluminum housing. This self-contained stereo-camera system is fitted to the aft end of a trawl, which is attached to a moving boat, in place of the codend (i.e., capture bag) for video sequences capturing. The absence of the codend allows fish to pass unharmed to the environment after being sampled (video captured).

The high-resolution high-sensitivity cameras are capable of capturing 4-megapixel images. The cameras are connected via a gigabit Ethernet to a Core 2 Duo PC with software to control the camera's operation and to store the video data to a solid state hard disk drive. Due to the limited bandwidth of Ethernet data transmission and storage in an earlier hardware design of the Cam-trawl, the capturing rate of cameras is at most 10 frames per second (fps). Considering the tradeoff between the image quality and data transmission speed, we set the capturing rate to 5 fps. This allows the cameras to collect high-definition video data that are favorable for accurate segmentation and tracking. At this capturing rate, targets move abruptly from one frame to another and enter/exit the field of view (FOV) frequently (4.3 frames of target lifespan in average). This makes conventional tracking methods infeasible for this task. To illustrate this scenario, six consecutive frames captured by the Cam-trawl are shown in Fig. 1 (b). A full-featured software development kit (SDK) supports the core acquisition and control routines. The PC runs a customized Linux operating system, which allows precise control over what software and services are started depending on how the system is being used.

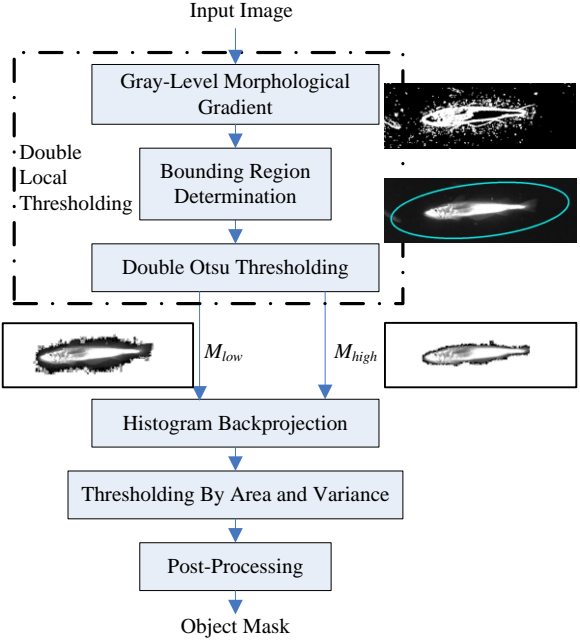

Fig. 2. Flow chart of the proposed fish segmentation algorithm.

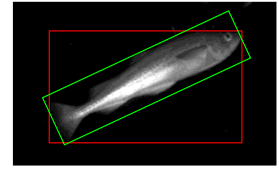

Fig. 3. Upright (red) and oriented (green) bounding box of a target.

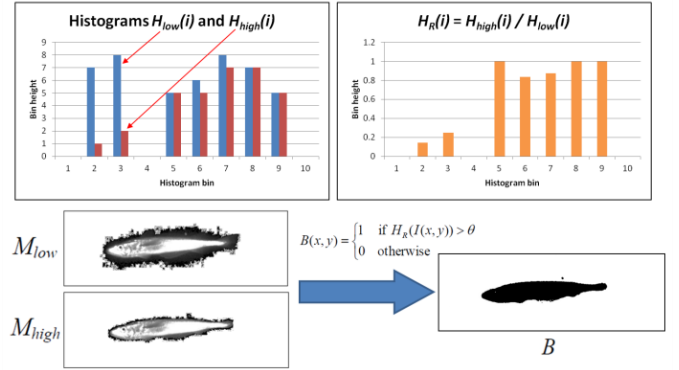

Fig. 4. Basic concept of histogram backprojection.

## III. AUTOMATIC FISH SEGMENTATION

The proposed algorithm for automatic fish segmentation is divided into four steps, as shown in Fig. 2. The first step is double local thresholding. Next, binary object masks generated by two different thresholds are effectively integrated using a method based on histogram backprojection. After that, thresholding by area and variance removes noise and unwanted objects. Finally, a post-processing step is applied in order to refine the segmented object boundaries. Double local thresholding is designed to resolve the problem of non-uniform illumination over the video frame by focusing only on the vicinity of each target. On the other hand, histogram backprojection effectively merges two segmentation masks intelligently according to their difference in intensity distribution and refines the segmentation boundary for low-contrast imaging at the same time.

For the convenience in describing each target and its neighborhood, we use two types of bounding boxes for the objects in the rest of this paper: 1) upright bounding box and 2) oriented bounding box. An upright bounding box is an axis-parallel rectangle that encloses the object. It is commonly used to indicate the region where an object appears in an image or video frame. On the other hand, an oriented bounding box is a rotated rectangle that encloses the object. Its width, height and orientation are determined by the principal component analysis (PCA) of the binary object mask. More specifically, the width is the object size measured along the direction of the first principal component, the height is that measured along the second principal component, and the orientation of the rotated box is parallel to the first principal component. Examples of upright and oriented bounding boxes are illustrated in Fig. 3.

### A. Double Local Thresholding

One approach to object segmentation for video with a simple background, as shown in Fig. 1 (b), is thresholding, which binarizes the video frame by setting a threshold on pixel intensity. Otsu's method [21] is widely used to find the optimal threshold, which separates the histogram into two classes so that the combined intra-class variance is minimal. Using only one threshold, however, introduces defects in object contours when the contrast between foreground and background is low. Also, thresholding over the entire video frame usually fails to segment objects if the illumination is uneven across the frame. To overcome these challenges, the double local thresholding algorithm is proposed to find two thresholds, i.e., generate two different binary masks, within each object's neighborhood. These two binary masks will be merged and refined by the subsequent histogram backprojection described in the next subsection.

When using double local thresholding for fish segmentation, we need to first detect a rough position and size of the fish. A gray-level morphological gradient operation [20] is performed on the input video frame to roughly locate the fish object in the input video frame. Next, the local region around the detected objects has to be determined. The classic connected components algorithm [22] is applied to mark the isolated local region in the object mask. Each region is then described by an inscribed ellipse of the oriented bounding box. The length of major and minor axis of the ellipse is the width and height of oriented bounding box, respectively. The orientation of ellipse is the same as the oriented rectangular box. Finally, the local region is determined by enlarging the oriented ellipse by a factor of 1.5 in both the major and minor axes.

With these elliptic local regions, we are ready to perform the double local thresholding method. For each region, adaptive thresholds are selected using our proposed variant of the Otsu's method. To preserve some dim targets, which have intensity values close to the background, the threshold is given by

$$\tau_x = \tau - p(\tau - \mu_L(\tau)) \,, \tag{1}$$

where $\tau$ is the threshold given by Otsu's method, $\mu_L(\tau)$ is the mean of lower class separated by $\tau$ in the histogram and $p$ is a shifting factor. Using (1), two thresholds are obtained by setting different values for $p$, i.e., $\tau_{low} = \tau - p_{low}(\tau - \mu_L(\tau))$ as the low threshold and $\tau_{high} = \tau - p_{high}(\tau - \mu_L(\tau))$ as the high threshold. Applying these two thresholds to the local region in the video frame results in two corresponding object masks $M_{low}$ and $M_{high}$, as shown in Fig. 2. A 3×3 median filter is applied to both binary object masks as in [28], [29] to reduce the impulsive noise before being merged by histogram backprojection described in the next subsection.

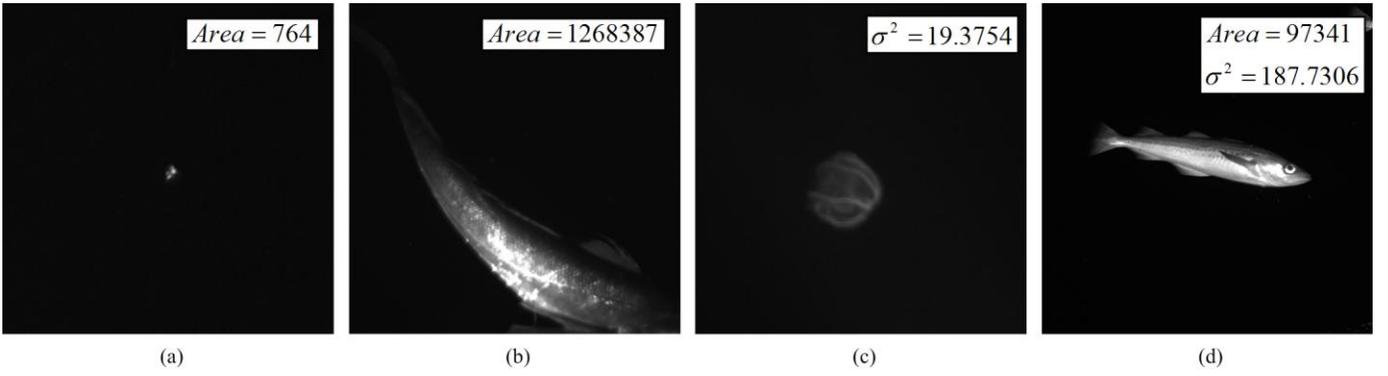

Fig. 5. Example objects that are discarded due to (a) small area, (b) large area and (c) small variance of pixel values; (d) an object that are preserved after thresholding by area and variance. Note that (a) and (b) present the entire video frame, while (c) and (d) are zoomed-in regions.

### B. Histogram Backprojection

In the underwater scenario, unstable lighting condition causes the low contrast in captured video data. As a result, the segmentation of fish is usually defective, especially around the boundary. It is thus desirable to refine the boundary of segmentation by comparing and aggregating two object masks according to their distributions of pixel values since they cover different amount of background pixels [14].

To check whether a specific pixel $I(x, y)$ within the bounding region of an object candidate belongs to the foreground or background, the histogram backprojection defined in (3) is adopted. First, two gray-level histograms $H_{low}(r)$ and $H_{high}(r)$ are computed according to the object masks $M_{low}$ and $M_{high}$, respectively. A ratio histogram of any gray-level value $r$ is defined as

$$H_R(r) = \min\left(\frac{H_{high}(r)}{H_{low}(r)}, 1\right). \tag{2}$$

Next, the ratio histogram is backprojected to the video frame domain, i.e., $BP(x, y) = H_R(I(x, y))$, $1 \le x \le W$, $1 \le y \le H$, where $I(x, y)$ denotes the pixel value at $(x, y)$. A thresholding process is then applied to the backprojection of the ratio histogram $H_R(r)$, and the final binary segmentation mask $B(x, y)$ is given by

$$B(x, y) = \begin{cases} 1 & \text{if } BP(x, y) > \theta_{bp} \\ 0 & \text{otherwise} \end{cases}, \tag{3}$$

where $\theta_{bp}$ denotes a threshold between 0 and 1. An illustrative example for the basic concept of histogram backprojection is shown in Fig. 4.

### C. Thresholding by Area and Variance

In addition to using histograms to refine the segmentation masks, the proposed algorithm also takes into account the area of an object and variance of pixel values within an object. The connected components algorithm is applied to determine each isolated blob with its area. Those objects whose areas are greater than an upper threshold (corresponding to targets which are too close to the cameras with partial fish body capturing) or less than a lower threshold (corresponding to noise or very far away fish which cannot be reliably measured) will be rejected. Specifically, for each pixel $(x, y)$ within the $k$-th segmented object $O_k$, its corresponding pixel on the foreground mask is revised by

$$B(x, y) = \begin{cases} 1 & \text{if } \theta_A^L \le A(O_k) \le \theta_A^U \\ 0 & \text{otherwise} \end{cases}, \quad (x, y) \in O_k, \tag{4}$$

where $A(\cdot)$ gives the area of an object, and $(\theta_A^L, \theta_A^U)$ are the lower and upper bound of the area to preserve.

Object candidates are also examined by calculating the variance of pixels within each segmented objects. Since foreground objects (fish) are inclined to be more textured than the background or unwanted objects, the variance of the segmented object is likely to be larger. The variance of pixels for an object is given by

$$\sigma_k^2 = \frac{1}{A(O_k)-1} \sum_{(x,y) \in O_k} \left( I(x,y) - \overline{I}_k \right)^2, \quad (x,y) \in O_k, \qquad (5)$$

where $\overline{I}_k$ denotes the mean of pixel values of the $k$-th object. Given the variance, the foreground mask for this object is then thresholded by

$$B(x,y) = \begin{cases} 1 & \text{if } \sigma_k^2 \geq \theta_V \\ 0 & \text{otherwise} \end{cases}, \qquad (6)$$

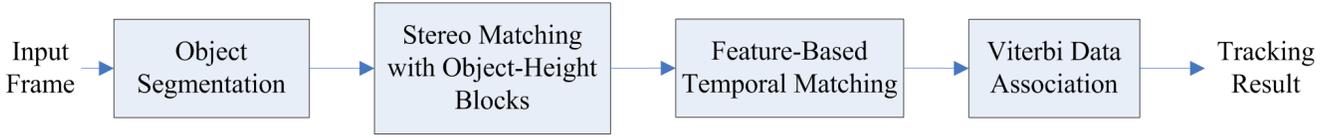

Fig. 6. Flow chart of the proposed multiple fish tracking system.

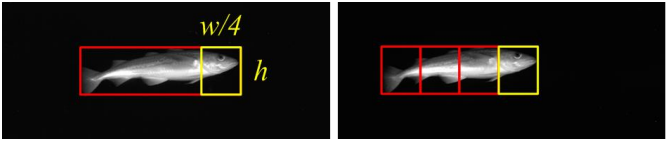

Fig. 7. Object-height blocks (yellow) on the stereo-rectified left and right image. There are 4 candidates from each of the right target's upright bounding box for the minimum-SAD scheme.

### D. Post-processing

There may still exist some errors, such as gulfs or peninsulas, created at the boundaries of histogram backprojection refined objects. A cascade of morphological operations can be adopted to further refine the segmentation boundaries. More specifically, a closing followed by an opening morphological operation with a disk structuring element is applied to the object mask. In the experiments, we empirically choose the size of the structuring element as $7 \times 7$ pixels. In this way, the object boundaries are smoothed without affecting the details of the shape information.

## IV. TRACKING UNDER LOW FRAME RATE

Most moving objects in the low-frame-rate video are much more difficult to track due to their poor motion continuity and frequent entrance/exit. This makes most standard video object tracking methods as [7], [8] infeasible. An overview of the proposed multiple fish tracking algorithm is shown in Fig. 6. Fish segmentation described in Section III is performed and followed by the proposed fast stereo matching method to match objects in two cameras. With the result of segmentation and stereo matching, fish are tracked by the feature-based temporal matching and the multi-target Viterbi data association. The proposed tracking algorithm exploits the temporal relationships throughout the target lifespan instead of between only the latest two frame. As a result, fish targets can be tracked in a more robust way even the motion is abrupt under the low frame rate.

### A. Stereo Matching with Object-Height Blocks

Stereo imaging allows for a mean to obtain depth information and validate the result of object tracking. However, one major drawback of traditional dense stereo matching techniques is the intensive computations. When tracking fish using the proposed approach in the following subsections, stereo vision techniques are utilized for pairing the left and right targets. Therefore, with knowledge of the fish target location available in the segmentation stage, an efficient block matching approach is proposed to find the stereo match for each target while reducing much of the computation.

Before stereo matching, preprocessing such as stereo calibration and stereo rectification are applied to each pair of input video frame. Through these steps, camera views are undistorted and transformed so that each pair of corresponding epipolar lines in both cameras are exactly horizontal and align with each other in terms of vertical location. Stereo matching is hence reduced to a 1-D search problem along a horizontal line. It also ensures that the height of an object is equal in two camera views. In light of this, we introduce the notion of object-height blocks to speed up the matching process.

Given a segmented object in the left video frame, its upright bounding box is equally divided to 4 non-overlapping blocks horizontally, as shown in Fig. 7. These blocks are referred to as *object-height blocks*. For each object-height block in the left video frame, the best match in the right video frame is determined by a simple block-matching algorithm based on the minimum sum of absolute difference (SAD) criterion. Note that an object-height block only searches along the horizontal line, and takes as candidates only those blocks within the upright bounding box of the object. That is, each target appearing on the horizontal line of the right video frame provides 4 candidates for a block from the left video frame to match. This results in great saving of computations with little loss of the accuracy, which is tolerable for the purpose of pairing objects in the left and right cameras. The object in the right view that has the minimum of the sum of 4 object-height blocks' SADs is then selected as the corresponding object of the left target.

### B. Feature-Based Temporal Matching

The ubiquitous noise introduced by organic debris and abrupt movement of targets are the major difficulties for tracking in a low-frame-rate underwater video. An object matching approach is therefore developed for associating observations and targets. Various useful features are considered for measuring the similarity between objects. Given objects $O_t^j$ in frame $t$ and $O_{t-1}^i$ in frame $(t-1)$, four cues are investigated as follows.

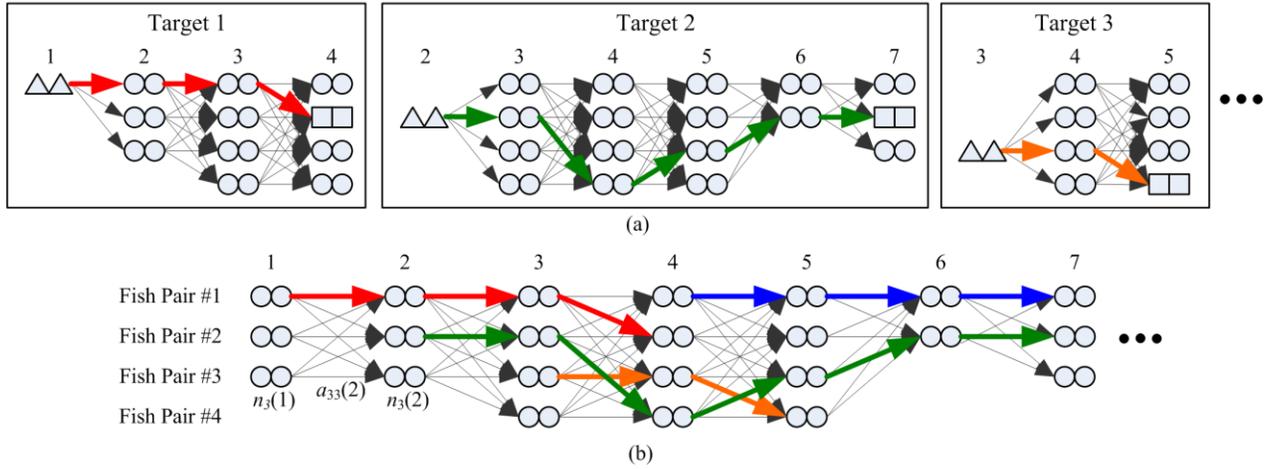

(a)

(b)

Fig. 8. Multiple-target Viterbi data association. (a) Each target maintains a separate trellis during its lifespan with its own starting node (triangles) and ending node (squares). The optimal path in each trellis is labeled by colored arrows. (b) An overall trellis showing several paths from targets in (a).

### 1) Vicinity cue

The Euclidean distance is given by $\left\| \mathbf{x}_t^j - \hat{\mathbf{x}}_t^i \right\|$, where $\mathbf{x}_t^j$ and $\hat{\mathbf{x}}_t^i$ denote the center point coordinates of the observation $j$ and the prediction of target $i$ at frame $t$, respectively. Details about target prediction obtained by a motion projection scheme are described in Section IV-C.

### 2) Area cue

In order to remove the factor of frame-by-frame discrepancy of target distance from cameras when calculating the area difference between objects, the object depth is calculated by stereo triangulation, and the target area is normalized accordingly as if all targets are placed at the same distance from the stereo cameras. The difference of area between the associated objects in two consecutive frames is supposed to be small. The object area, denoted as $A(\cdot)$, is calculated by the connected components algorithm. The difference of area between the object $O_t^j$ in frame $t$ and the object $O_{t-1}^i$ in frame $(t-1)$ is then given by $\left| A(O_t^j) - A(O_{t-1}^i) \right|$.

### 3) Motion direction cue

Given two objects $O_t^j$ and $O_{t-1}^i$ to be matched, we define the corresponding motion vector as $\mathbf{v}_t^{i,j} = \mathbf{x}_t^j - \mathbf{x}_{t-1}^i$. The direction of motion is then represented by the angle $\theta(\mathbf{v}_t^{i,j}, \mathbf{v}_{ref})$ between $\mathbf{v}_t^{i,j}$ and a predefined reference vector $\mathbf{v}_{ref}$, given by

$$\theta(\mathbf{v}_t^{i,j}, \mathbf{v}_{ref}) = \cos^{-1} \frac{\mathbf{v}_t^{i,j} \cdot \mathbf{v}_{ref}}{\left\| \mathbf{v}_t^{i,j} \right\| \left\| \mathbf{v}_{ref} \right\|}. \tag{7}$$

The predefined reference vector can be chosen according to the motion trend of fish schools or the movement of cameras.

### 4) Histogram distance

In addition to geometric features, the appearance (pixel intensity) also plays an important role. To exploit the dissimilarity of intensity distribution between two objects, the earth mover's distance (EMD) [24] is computed as the distance metric between 16-bin gray-level histograms.

Combining all the four cues above, the likelihood for object temporal matching between $O_t^j$ and $O_{t-1}^i$ is given by

$$
\begin{aligned}
P(O_{t-1}^i \mid O_t^j) &= \exp(-\frac{\left\| \mathbf{x}_t^j - \hat{\mathbf{x}}_t^i \right\|^2}{\sigma_v^2}) \cdot \exp(-\frac{(A(O_t^j) - A(O_{t-1}^i))^2}{\sigma_a^2}) \\
&\quad \cdot \exp(-\frac{(\theta(\mathbf{v}_t^{i,j}, \mathbf{v}_{ref}))^2}{\sigma_m^2}) \cdot \exp(-\frac{(EMD(O_t^j, O_{t-1}^i))^2}{\sigma_h^2}) \\
&= \exp(-z_v) \cdot \exp(-z_a) \cdot \exp(-z_m) \cdot \exp(-z_h) \\
&= \exp(-(z_v + z_a + z_m + z_h)),
\end{aligned}
\tag{8}
$$

and the "matching cost" is defined as

$$c_{ij}(t) = -\ln P(O_{t-1}^i \mid O_t^j) = z_v + z_a + z_m + z_h. \tag{9}$$

The $\{\sigma\}$ values in (8) denote the features' standard deviations. These standard deviations are calculated systematically by collecting the feature values for all temporal matching candidates in each frame of our video data. This cost is assigned to the edge between $O_{t-1}^i$ and $O_t^j$ in the trellis for Viterbi data association.

### C. Viterbi Data Association

In the proposed data association system, the stereo information is utilized by regarding a pair of stereo-matched object, i.e., the same object in the left and right video frame, as one observation for tracking. The matching cost of temporal matching is then given by the sum of the costs from two cameras, i.e., $c_{ij}^{stereo}(t) = c_{ij}^L(t) + c_{ij}^R(t)$. The advantage of combining a stereo pair of objects can be seen from two aspects. First, the stereo-matched objects are ensured to be bound together during

---

**Algorithm 1.** Viterbi Data Association At Each Frame

---

1.   **input:** time $t$ , objects $\{O_i^j\}_{j=1}^{|N(t)|}$ , trellises $\{\mathbf{T}^k\}_{k=1}^m$

2.   **output:** target paths $P(t)$ ending at time $t$

3.   initialize the object selection set $S \leftarrow \varnothing$

4.   **for** $k = 1$ to $m$ **do**

5.       estimate $c_{ij}(t)$ for all $i, j$ based on (9)

6.       update $\mathbf{T}^k$ with $\pi_j(t), C_j(t)$ based on (11), (12)

7.       $S \leftarrow S \cup \arg\min_j C_j^{stereo}(t)$

8.       **if** target $k$ exits the FOV **then**

9.          $P(t) \leftarrow P(t) \cup Backtracking(\mathbf{T}^k)$

10.      **end if**

11.   **end for**

12.   create a new trellis $\mathbf{T}^j$ for each $O_i^j \notin S$

---

tracking, so the mismatch between tracking and stereo vision is greatly reduced. Also, an object pair is considered the optimal candidate for a tracked target only if this object pair matches well in both the left and right images. In other words, pairing of objects with stereo cameras enables an implicit tracking validation, which is not available in the single camera scenario. This is especially helpful for improving tracking performance in our scenario, where targets to be tracked can be visually similar.

To exploit temporal correlations of objects across several past frames, a multiple-target Viterbi data association algorithm is introduced. Algorithm 1 summarizes the procedure for each video frame.

### 1) Basic idea

The Viterbi data association [17]–[19] is performed based on a trellis of the observations in each frame. As illustrated in Fig. 8, a trellis is a type of directed graph in which nodes are partitioned into ordered subsets $N(t) = \{n_j(t) \mid j = 1, 2, ..., |N(t)|\}$ for $t = 1, 2, ..., T$ , and edges $a_{ij}(t)$ lie between any pair of node in adjacent subsets $\{n_i(t-1), n_j(t)\}$ . Nodes in a subset represent objects in one frame, and each edge is assigned a cost $c_{ij}(t)$ . The total cost of a path (a sequence of edges) is then given by

$$C(P) = \sum_{t=2}^{T} c_{p_{t-1}p_t}(t) ,$$

$$\text{where } P = \{a_{p_1p_2}(2), a_{p_2p_3}(3), ..., a_{p_{T-1}p_T}(T)\} . \quad (10)$$

The Viterbi algorithm [25] is applied here to find the minimum-cost path during single-target tracking. For every observation a node is initialized with zero cost and a null predecessor. In each iteration, the matching cost for every node $n_j(t), j = 1, 2, ..., |N(t)|$ is given by (9). Then the predecessor and accumulated cost are assigned to node $n_j(t)$ :

$$\pi_j(t) = \arg\min_{1 \le i \le |N(t-1)|} C_i(t-1) + c_{ij}(t) . \quad (11)$$

$$C_j(t) = C_{\pi_j(t)}(t-1) + c_{\pi_j(t)j}(t) . \quad (12)$$

Once the target leaves the FOV (as established by distance from video frame borders, see below), i.e., the final stage of trellis is reached, a backtracking step is performed. Starting from the minimum-cost node in the final stage, the optimal path $P^* = \arg\min_P C(P)$ is recovered by traversing backward to the first stage according to the predecessors stored at each stage.

### 2) Multiple-Target Case

A multiple-target Viterbi data association algorithm is proposed for our case of low-frame-rate fish tracking [16]. Since the starting frame may differ among targets, the predecessor and minimum cost at each node may also differ. Therefore, we create a separate trellis for every target to track. Data association is then performed separately by following (11) and (12) for each target with all observations, as shown in Fig. 8. Observations that are not associated with any target corresponds to new targets or false alarms. New targets tend to appear close to the video frame border in their first frame. Also, most false alarms here are generated during the post-processing of segmentation, so their areas are usually small. An examination on position and area of the observations is thus performed to distinguish new targets from false alarms. For exiting targets, the abrupt movement and short lifespan make the criteria of track creation and deletion less reliable. Therefore, a track is restricted to end only when its predicted position is close to the frame border. This also prevents targets from being deleted if they are temporarily occluded. If a track is lost before approaching the frame boundary, the predicted position is used as the actual position and the velocity remains. A new prediction is then made for the next frame.

Note that occlusions are inherently handled by our method since paths in different trellises may share the same nodes. In terms of the overall trellis in Fig. 8 (b), nodes in any stage in the diagram are allowed to be included by more than one path, which means the object in this frame is occluded by others.

In each frame, a motion projection mechanism is utilized to estimate and update the position of the tracked target. Given the position $\mathbf{x}_{t-1}^k$ and velocity $\mathbf{v}_{t-1}^k$ of the $k$ -th tracked target from frame $(t-1)$ , the predicted position at the current frame is given by $\hat{\mathbf{x}}_t^k = \mathbf{x}_{t-1}^k + \mathbf{v}_{t-1}^k$ . After the data association, the observation node with the minimum cost is chosen to update the position and velocity by

$$\mathbf{x}_t^k = \mathbf{x}_t^{j*}, \quad (13)$$

$$\mathbf{v}_t^k = \alpha \mathbf{v}_t^{j*} + (1-\alpha)\mathbf{v}_{t-1}^k, \quad (14)$$

where $\mathbf{x}_t^{j*}$ and $\mathbf{v}_t^{j*}$ denote separately the position and velocity of the minimum-cost observation and $\alpha$ is the update rate.

Rather than using a batch scheme as in [17], which accumulates a fixed number of frames and perform backtracking periodically, in the proposed scheme backtracking is performed and the optimal sequence of observations is recovered once there is a target leaving the FOV. The proposed system is thus able to not only perform online tracking but also exempt from potential failures due to the gaps between batches.

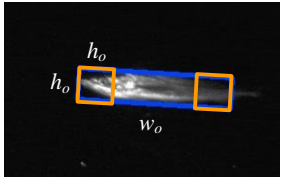

Fig. 9. End square regions (orange) of a fish, where $w_O$ and $h_O$ denote the width and height of the oriented bounding box (blue).

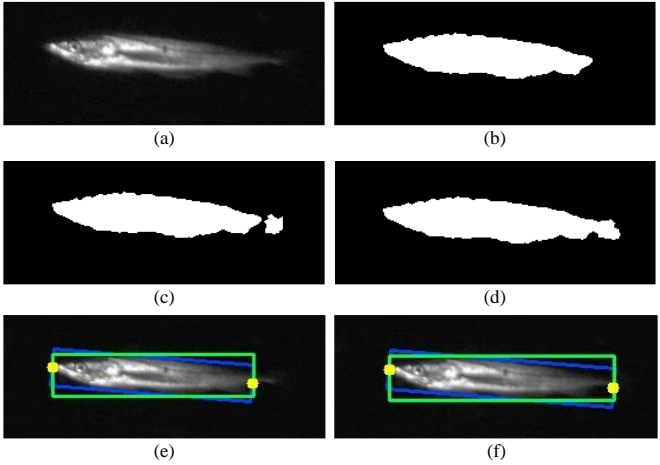

Fig. 10. Illustrative example of fish-tail end compensation: (a) stereo-rectified image; (b) object mask before end compensation; (c) result of snake algorithm; (d) new object mask after morphological closing operation; (e) upright bounding box (green), oriented bounding box (blue) and end points (yellow) before end compensation; (f) upright bounding box, oriented bounding box and end points after end compensation, where the less-reflective part around tail fin is recovered.

### D. Fish Length Measurement in 3-D Space

In addition to tracking and counting, another beneficial information for fish abundance survey comes from knowing the length estimate of each fish in the video. By capturing fish video with the Cam-trawl, length measurement of live fish can be done by employing stereo imaging techniques. Stereo matching guided by the object-height blocks, as described in Section IV.A, matches fish targets between the left and the right video frames and generates a coarse disparity map. To enhance the accuracy of fish length measurement, a fish-tail end compensation step is further applied to each pair of stereo-corresponding targets. Disparity refinement followed by stereo triangulation is performed to estimate the 3-D length of fish body.

#### 1) Fish-Tail End Compensation

A common failure case in fish segmentation [15] is fish cropping caused by the lower reflectivity of the caudal fin (tail). This introduces considerable error in fish length measurement. To overcome this issue, a fish-tail end compensation technique using the result of stereo matching is performed.

To obtain a region for the possible presence of fish tail, we utilize the oriented bounding box of fish introduced in Section III. The end region of a fish is defined as a square with size

#### TABLE I
#### VALUES OF PARAMETERS USED IN THE EXPERIMENTS

| Symbol | Value | Description |
|---|---|---|
| *Automatic Fish Segmentation* | | |
| $p_{low}$ | 1 | Shifting factor of low threshold |
| $p_{high}$ | 0.7 | Shifting factor of high threshold |
| $\theta_{bp}$ | 0.3 | Threshold of histogram backprojection |
| $\theta_A^L$ | $2 \times 10^3$ | Lower limit of object area |
| $\theta_A^U$ | $10^6$ | Upper limit of object area |
| $\theta_V$ | 30 | Threshold of object variance |
| *Multiple-Fish Tracking* | | |
| $\mathbf{v}_{ref}$ | $(-1,0)$ | Reference motion vector |
| $\alpha$ | 0.3 | Update rate for target velocity |
| $M$ | 100 | Margin from frame border for target entrance/exit |
| *Fish Length Measurement* | | |
| $\theta_{SAD}$ | 16 | Threshold of SAD for body end mismatch |
| $r$ | 16 | Search range for disparity refinement |

$h_0 \times h_0$, where $h_0$ is the height of oriented bounding box. An example of end square region is illustrated in Fig. 9. For a pair of stereo-matched objects, the two ends of the objects are validated by computing the SAD of two end regions between the left and the right video frames. A mismatch is detected via thresholding the ratio of the computed SAD value to the average minimum SAD of the same target. More specifically,

$$mismatch = \begin{cases} 1 & \text{if } SAD/\mu_{SAD} > \theta_{SAD} \\ 0 & \text{if } SAD/\mu_{SAD} \le \theta_{SAD} \end{cases}, \quad (15)$$

where $\mu_{SAD}$ is the average of the minimum SAD values of the four object-height blocks associated with the same target, and $\theta_{SAD}$ is an empirically defined threshold value.

End compensation method is applied once a mismatch is detected. The classic "Snake" (deformable contours) algorithm [23] is used to extract the missing tail. We use a circle with a diameter of $h_0/2$ as the initial "Snake" and place it next to the mismatch end along the principal component of the target. After the iterations, the extracted region is connected to the body by using a morphological closing operation. The effect of the fish-tail end compensation procedure is demonstrated in Fig. 10.

#### 2) Body Length Estimates

Measuring the target length requires more accurate information of depth. A disparity refinement based on coarse disparity map is therefore applied. Same as in Section IV.A, a block-matching approach with SAD criterion is used. To measure the fine disparity, we use dense grid blocks for matching and set the search range as $[d_o - r, d_o + r]$, where $d_o$ is the coarse disparity at that position as derived from stereo matching based on object-height blocks, and $r$ denotes the search range size.

| Num. of Targets | Precision | Recall |
|---|---|---|
| 514 | 0.746 | 0.784 |

TABLE III
MEAN ABSOLUTE PERCENTAGE ERROR OF LARGE TARGET LENGTH

| Num. of Large Targets | MAPE of Length |
|---|---|
| 189 | 10.7 |

TABLE IV
PRECISION AND RECALL OF FISH DETECTION OVER VIDEO CLIPS

| Num. of Targets | Precision | Recall |
|---|---|---|
| 62 | 0.98 | 0.94 |

TABLE V
TRACKING SUCCESS RATE VS. DATA ASSOCIATION METHODS

| Clip | NN | Shape | [27] | MC | MC+VDA |
|---|---|---|---|---|---|
| 1 | 0.38 | 0.31 | 0.66 | 0.47 | 0.94 |
| 2 | 0.42 | 0.50 | 0.58 | 0.58 | 0.83 |
| 3 | 0.29 | 0.44 | 0.44 | 0.50 | 0.86 |
| **Avg.** | 0.36 | 0.41 | 0.56 | 0.54 | **0.88** |

The technique of stereo vision leads the way to absolute length measurement of targets by locating the targets back in the 3-D space. Intrinsic parameters of the cameras have been obtained in prior by stereo calibration. Given a point on the 2-D image represented by its screen coordinates $(x, y)$ and disparity $d$, it can be projected into 3-D space by using stereo triangulation [20]. The length of the $i$-th fish is thus given by $L(O_i) = \left\| \mathbf{x}_i^H - \mathbf{x}_i^T \right\|_2$, i.e., the Euclidean distance between the head point and tail point of the fish body in the 3-D space.

## V. EXPERIMENTAL RESULTS

### A. Data and Implementation Settings

Simulations of the proposed system are carried out on several 8-bit grayscale video clips recorded underwater by the stereo cameras on Cam-trawl. The frame size is $2048 \times 2048$ pixels, and the frame rate is 5 frames per second. Before the object segmentation stage, a pre-processing is performed to eliminate the trawl web in video frames. The trawl web behind the targets appears as white diagonal grids spreading over the top and bottom regions in the stereo camera views (see Fig. 11). Its drift with the current makes background modeling approaches ineffective in this scenario. Observing its shape, we remove the trawl web successfully by applying morphological opening operations with main diagonal (from top-left to bottom-right) and anti-diagonal (from top-right to bottom-left) structuring elements with length of 7 pixels. For the rest of the proposed algorithm, the structuring element for all morphological operations is empirically determined as a disk with size $7 \times 7$ pixels. Values of parameters that are determined empirically in our experiments are provided in Table I. The whole process is fully automatic and requires no manual intervention. Tracked fish are labeled with numbers and bounding boxes with different colors in order to make them differentiable.

### B. Result of Fish Segmentation

The proposed segmentation algorithm is tested with three sample video sequences consisting of 74 frames in total to evaluate its performance. According to the hand-labeled ground truth, there are 514 fish in total to be segmented. The fish length is defined as the Euclidean distance between head and tail. Note that fish lengths are estimated only from a single camera in this experiment. In the next subsection, fish lengths measured from stereo cameras will be discussed and compared with the single-camera case.

The performance of fish segmentation is measured in terms of precision and recall accuracy as well as the mean absolute percentage error (MAPE) of the measured length of "large targets", which are the ones with length greater than 100 pixels, since they have more reliable ground truth. There are 189 large targets out of 514 targets in the testing set. In this experiment, the length of a fish object is estimated by finding its oriented bounding box. The measured fish length equals to the maximum between the width and height of the oriented bounding box.

As shown in Table II, the proposed algorithm achieves a 74.6% precision and a 78.4% recall under very low-contrast underwater videos. The MAPE of measured length out of 189 large targets is 10.7%, as shown in Table III. This shows that the proposed segmentation algorithm gives quite accurate information of fish silhouette for the use of stereo matching as well as target tracking. A major source of error is the fish cropping that happens often because of the low reflectivity of caudal fins. Based on this, the performance of length estimation and the object detection is further enhanced by the stage of stereo matching and multiple-fish tracking, respectively, as described in the following subsections.

### C. Result of Multiple Fish Tracking

The proposed system is used to track multiple fish targets simultaneously in several sample video clips. All these video clips are grayscale and recorded underwater by the stereo cameras on Cam-trawl. Some tracking statistics of the proposed system comparing with other data association methods are listed in this subsection.

From Table IV, the precision and recall of fish detection in sample video clips containing 62 distinct fish targets are 98% and 94%, which shows that the proposed system enhances greatly the accuracy of underwater fish detection by utilizing temporal information. In Table V, the performance of the proposed system is evaluated and compared with other data association algorithms. Here, the tracking success rate is defined as the ratio of correctly tracked targets to correctly detected targets, i.e.,

$$\text{tracking success rate} = \frac{\text{\# of targets correctly labeled}}{\text{\# of targets correctly detected}}. \quad (16)$$

One can see that the proposed system, i.e., matching cost plus Viterbi data association (MC+VDA) outperforms other data association methods, including the state of the art [27]. The conventional nearest neighbor (NN) suffers from poor motion continuity and short lifespan of targets, and thus tracks fish in a low success rate. The shape-based method (Shape) finds the candidate with the most similar shape to the target in each frame. In the experiment, we use the Hu's invariant moments [26], which are known for their robustness against image translation, rotation and scaling. Shape-based method thus performs slightly better than the nearest neighbor approach. However, unlike pedestrians or vehicles, fish in an unconstrained environment are considered deformable bodies since their poses change abruptly during common actions such as swimming and turning. This is obvious especially when the video is captured at low frame rate, where the time interval between two consecutive frames is long (0.2 seconds in our data). For this reason, shape does not serve as a good feature to track fish in this case.

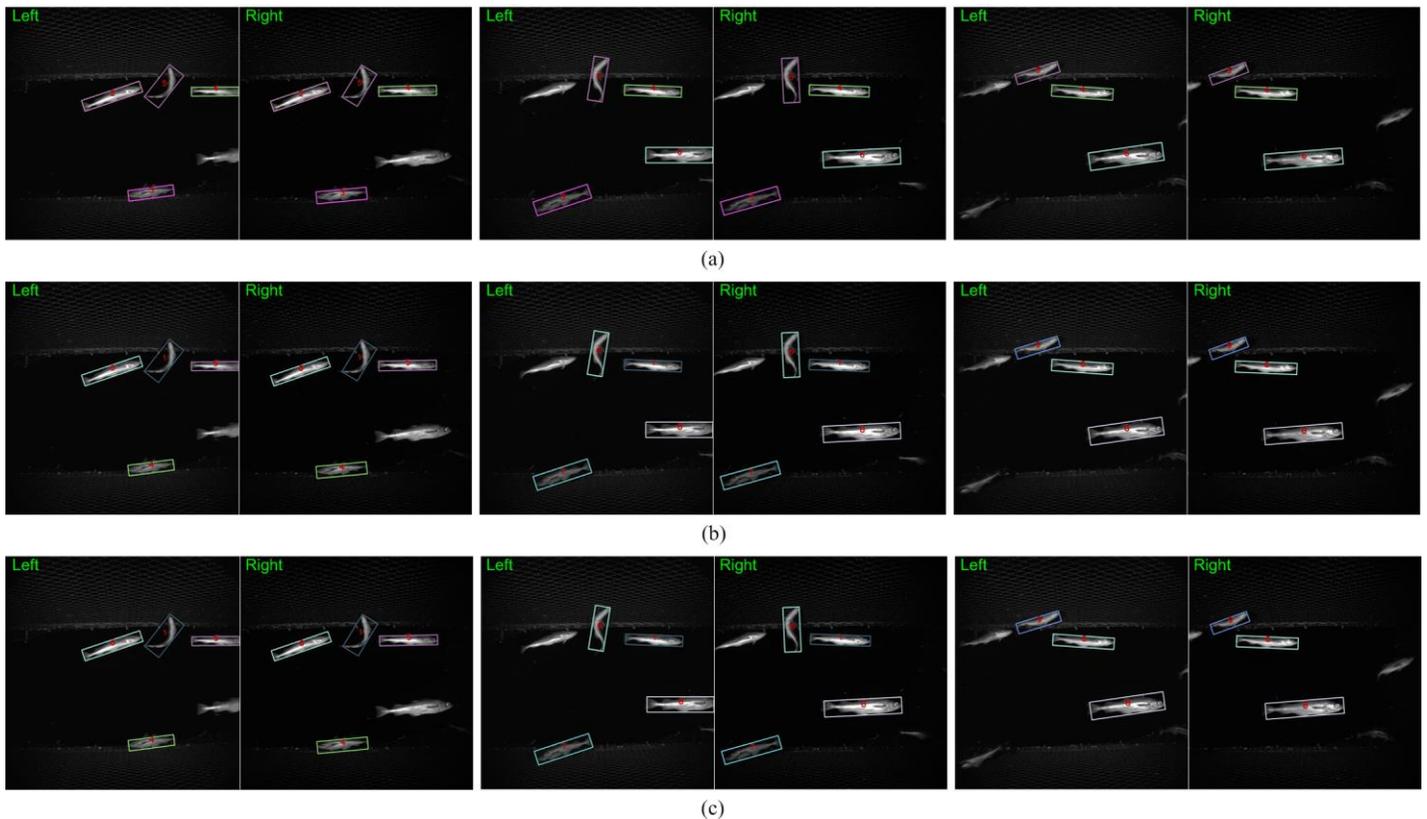

Fig. 11. Tracking multiple fish in an underwater video clip. Targets are labeled by numbers and oriented bounding boxes with different colors. (a) Using the proposed algorithm with matching cost plus Viterbi data association. (b) Using matching cost only. (c) Using primitive nearest neighbor data association.

The other compared method proposed in [27] exploits several features, including position, motion vector, area and orientation, which are similar to those used in the proposed method. However, same as conventional tracking algorithms, the temporal correlation of targets is only considered over two frames. This makes it less successful in the low-frame-rate case, where the motion can be very abrupt. Note that the proposed matching cost (MC) is also tested alone to show the effectiveness of matching objects based on various types of features. Without the Viterbi data association, the object matching scheme itself gives a tracking success rate close to that of [27]. By incorporating the Viterbi data association (VDA), our proposed method shows a much improved success rate for multiple target tracking.

The qualitative result in 6 consecutive frames with tracked fish labeled on both sides of stereo video frames is shown in Fig. 11, where comparative results from two other data association methods are also displayed. It clearly demonstrates the robustness of the proposed system. The stereo videos and our tracking results based on the oriented rectangular bounding boxes, along with the ground truth, are available at URL http://students.washington.edu/mengche/data/tracking, so that people can further develop useful algorithms to improve the performance.

### D. 3-D Fish Length Estimation

In addition to segmentation and tracking, we also evaluate the accuracy of fish body length measurement given by the proposed system. The testing set consists of video frames for one haul of fish consisting of 7120 frame. According to manually-measured ground truth, there are 316 targets in total captured in this haul by the Cam-trawl system. For disparity refinement, we use a dense block grid with size $8 \times 8$ pixels.

Thanks to the stereo matching and fish-tail end compensation, we are able to achieve 6.0% of mean absolute percentage error (MAPE) in fish length measurement, which greatly improves the previous measurements done with a single camera as in Table III. To further demonstrate the performance of the proposed algorithm, the length distribution of targets is also generated from the proposed algorithm and compared with two sources of ground truth. One is the physical measurements of the fish onboard, i.e., manual measurement of each individual fish with a ruler. The other is the measurements using a software program requiring manual selection of head and tail points for stereo-matched fish targets in the left and the right video frames.

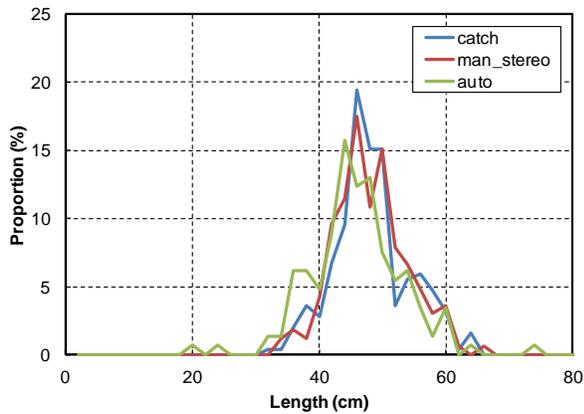

Fig. 12. Length frequency of 316 fish targets according to physical measurements (*catch*), 208 fish targets according to measurements using manual selection of end points (*man_stereo*) and using the proposed algorithm (*auto*). The decrease in target number from manual to automatic count is due to the low frame rate of video capturing so that a number of targets never appear completely inside the field of view in any frame.

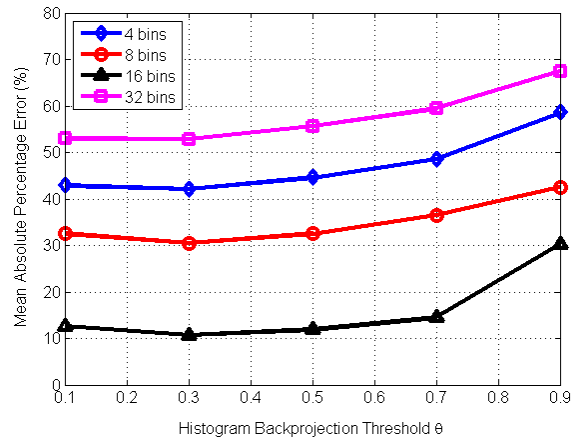

Fig. 13. Sensitivity test of histogram backprojection. Each color shows the error rates by using different numbers of histogram bins. Each labeled point shows the error rate by setting different values for the ratio histogram threshold.

As shown in Fig. 12, the proposed algorithm generates a good match with the two sources of ground truth in terms of the distribution of fish length frequency. The proposed algorithm gives slightly underestimated measurements because many fish in the video set have a certain extent of body bending, which makes them appear shorter. One way to compensate for this defect is a rescaling or regression technique on the data. A more precise approach, which will be one object of future work, is to establish a 3-D model for the fish body from the stereo cameras. By calculating the arc length of model midline, the length of fish body can be estimated with higher accuracy.

## VI. DISCUSSIONS

### A. Histogram Backprojection

As discussed in Section III.B, the contrast between the objects and the background is very low in underwater video data due to unstable lighting condition. This results in defective fish segmentation around the boundary. The boundary of the segmentation is successfully refined by utilizing the statistical distributions of pixel intensities in the neighborhood of an object. From the proposed double local thresholding, the low and high object masks have different sizes. Specifically, the low-threshold mask has a larger segmentation area, so it covers the foreground pixels better near the object boundary but meanwhile includes more background pixels. The high-threshold mask has a smaller segmentation area than the low mask, so it avoids covering background pixels but might lose a small portion of the foreground. Such differences in foreground and background pixel coverage are reflected in the mask histograms $H_{low}(r)$ and $H_{high}(r)$, where $H_{low}(r)$ consists of larger histogram values in the small gray-level bins (corresponding to background pixels) while $H_{high}(r)$ consists of lower histogram values in the small gray-level bins, as shown in Fig. 4 and Fig. 14 (c). The background pixels take a greater proportion in the low mask, so the bin height ratio $H_{high}(r) / H_{low}(r)$ is small for bins representing the background pixels. On the other hand, the foreground pixels take similar or equal proportion in two masks, so the bin height ratio $H_{high}(r) / H_{low}(r)$ is close to 1 for bins representing the foreground. Therefore, the ratio histogram $H_R(r)$ defined in (2) can be viewed as a confidence level of the pixel belongs to the foreground, and its backprojection provides a convenient way to examine each pixel around the boundary of the segmented object.

The effectiveness of segmentation boundary refinement based on histogram backprojection depends on the number of bins in each histogram as well as the thresholding value $\theta_{bp}$ for the ratio histogram in (3). To better understand the impacts of these parameters, we compared the performance of segmentation using 4, 8, 16 and 32 bins for each histogram with threshold values as 0.1, 0.3, 0.5, 0.7 and 0.9. Following the experiments in Section V.B, the performance is measured by the mean absolute percentage error (MAPE) of length obtained by segmentation for 189 large fish targets.

A sensitivity test on histogram bins and thresholds is reported in Fig. 13. One can see that using 16-bin histograms gives the lowest error in fish length estimation. The main reason is that a discriminative and robust representation for object appearance is crucial in low contrast and noisy underwater imaging. A histogram with fewer bins (e.g. 8 bins) mixes pixel values together and reduces the discrimination; a histogram with more bins (e.g. 32 bins) enhances the accuracy but become highly sensitive to noise. The 16-bin histogram groups similar pixel values appropriately, so it represents the object in a way which is not only discriminative but also robust against noise.

Moreover, one can see from Fig. 13 that the segmentation performance is rather insensitive to the threshold value $\theta_{bp}$ of ratio histogram. The reason can be seen from an example using

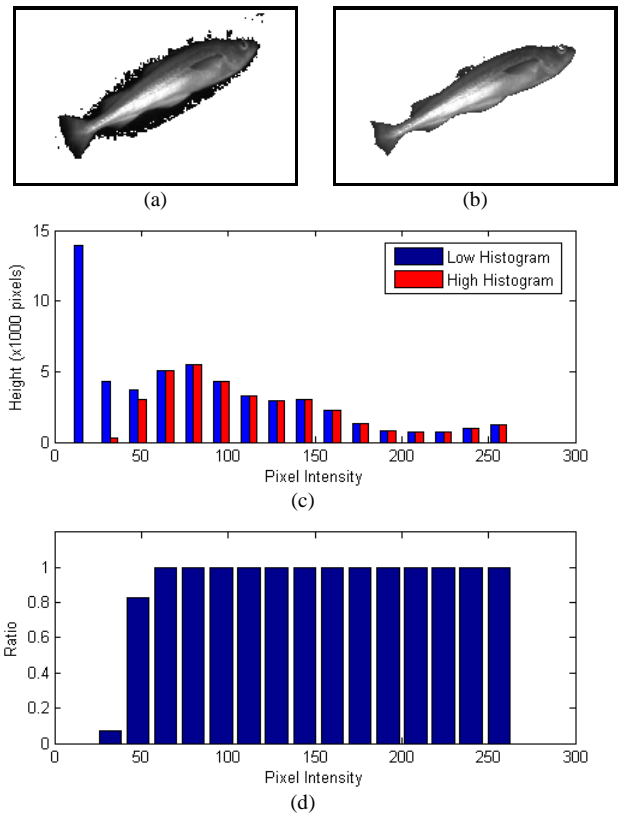

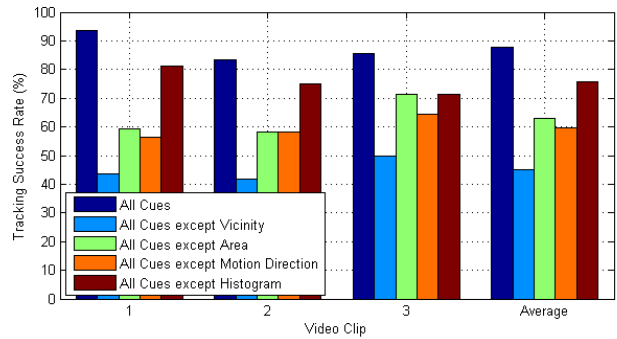

Fig. 15. Tracking success rate vs. unselected cues for temporal matching. Each color bar represents one cue that is not used when matching objects across video frames.

Fig. 14. An example target and its histograms generated by the proposed double local thresholding: (a) Low mask of the target; (b) high mask of the target; (c) two histogram from low and high mask, respectively; (d) ratio histogram given by (2). One can see that most ratio histogram bins are close to either 0 or 1. As a result, the performance of segmentation is less sensitive to the selection of ratio histogram threshold.

16 bins shown in Fig. 14. After our double local thresholding approach, almost all pixels covered by the high mask are also covered by the low mask. This gives the corresponding ratio histogram bins with heights being close to 1. On the other hand, the low mask covers much more dark background pixels than the high mask does (see the two leftmost bins in Fig. 14 (c)), so the heights of corresponding ratio histogram bins are close to 0. As a result, there is little difference in the performance as long as the threshold value is around the middle of the interval [0, 1].

### B. Features for Temporal Object Matching

The major challenge to tracking in low frame rate video is abrupt motion and short lifespan of targets. Traditional object tracking approaches such as particle filtering and kernel tracking fail to track in this case since they rely on the assumption of smoothness in motion trajectory, i.e., small incremental motion per frame. As a result, we resort to an approach related to object matching across frames.

In the proposed object temporal matching method, four different features are considered for measuring the similarity between objects in the current and previous frame. In order to further investigate the impact on tracking performance from each feature, a sensitivity test on these cues is carried out as follows. Using the same video data for evaluating tracking performance, one cue is removed from the matching cost function defined in (9) each time. The results of temporal object matching using partial cues are used by the subsequent multiple-target Viterbi data association. Finally, tracking success rate given in (16) for each video clip is calculated as a performance metric of multiple-target tracking. In this evaluation, we expect that matching objects with all cues gives the best performance, and discarding one of the cues results in a decreased performance. The importance of each removed cue is then reflected by the amount of drop in tracking success rate in the low frame rate video sequences.

Figure 15 reports the tracking performance with different removed cues in the matching cost function. An average success rate for each unselected cue is also given in addition to that for three testing video clips. As one can see from Fig. 15, the vicinity cue has the greatest impact on tracking performance, followed by the motion direction cue. This implies that the characteristics of motion, in spite of the poor continuity, is still informative to associate objects across frames. The concept of motion direction cue represents the directional property of not only the motion vector between two consecutive frames, but also the overall trajectory of a target throughout its lifespan. This is fully exploited by the Viterbi data association method since the dynamic programming approach provides a global optimization along the time horizon.

### C. Target Occlusions

Fish swimming through the nearly-unconstrained trawl often occlude each other in the underwater video. The low contrast imaging leads to fuzzy edges between two fish bodies when one fish overlaps the other. It is therefore difficult to deal with occlusions in fish segmentation. As discussed in Section IV.C, occlusions are inherently handled by our method by allowing paths in different trellises to share the same nodes. However, overlapping fish are viewed as one objects instead of separated targets during data association. One may thus be interested in whether the performance can be improved if occlusions are handled according to object depth before tracking.

To investigate into this, we perform the disparity refinement used for fish length measurement before tracking. Based on the coarse disparity in Section IV.A, the disparity is refined by using a dense grid of blocks with size $8 \times 8$ pixels. Block

| Method | Tracking Success Rate | MAPE of Length |
|---|---|---|
| w/o occlusion handling | 0.94 | 0.06 |
| w/ occlusion handling | 0.94 | 0.07 |

matching approach with maximum-SAD criterion is employed, same as in coarse matching. A simple Canny edge detection is then applied to the Gaussian-smoothed fine disparity map to separate objects when they are occluded.

For simplicity, video clip 3 is tested to see the influence of target occlusions. Results on fish tracking and length measurement are shown in Table VI. The success rate in tracking is the same with or without performing occlusion handling. This is expected since the proposed Viterbi data association allows paths from different trellises to intersect at some frame and even overlap for several frames. These correspond to one-frame and multiple-frame occlusion, respectively. As for length measurement, the MAPE is even higher after introducing occlusion handling. Despite the errors in measurements, this is caused by the fact that occluded fish are seriously underestimated in length after separation, even when the lengths of those fish which occlude others are measured more accurately.

## VII. CONCLUSION

A novel multiple fish tracking system is proposed for low-contrast and low-frame-rate underwater stereo cameras. Double local thresholding is developed to overcome the challenges posed by unstable illumination and ubiquitous noise in underwater imaging. Using histogram backprojection, we successfully generate a reliable fish segmentation in shape boundary under very low contrast. For low-frame-rate tracking, exploiting various appearance features, the cost function for feature-based object matching acts as an effective metric to find the temporal relationship of targets in the noisy underwater environment. Multiple-target Viterbi data association exploits multiple video frames from the past and takes advantage of dynamic programming to overcome the difficulties of abrupt target motion and frequent entrance/exit due to the low frame rate. Experimental result shows that the proposed system gives a success rate at 88% in terms of fish tracking for low-contrast and low-frame-rate underwater stereo videos. In addition, fish-body tail compensation enabled by stereo matching gives us 6% of mean absolute percentage error in fish length measurement under the low-contrast environment.